\documentclass{article}

\PassOptionsToPackage{numbers, compress}{natbib}

\usepackage[dblblindworkshop,final]{neurips_2025}

\workshoptitle{MATH-AI}

\usepackage[utf8]{inputenc} 
\usepackage[T1]{fontenc}    
\usepackage{hyperref}       
\usepackage{url}            
\usepackage{booktabs}       
\usepackage{amsfonts}       
\usepackage{nicefrac}       
\usepackage{microtype}      
\usepackage{xcolor}         
\usepackage{graphicx}
\usepackage{subcaption}
\usepackage{amsmath}
\usepackage{siunitx}
\usepackage{float}
\usepackage{placeins}  
\usepackage{wrapfig}

\title{Towards Scaling Laws for Symbolic Regression}

\author{
  David Otte$^{1}$ \quad
  Jörg K.H. Franke$^{1,2,3,4}$ \quad
  Arb{\"e}r Zela$^{2}$ \quad
  Fábio Ferreira$^{1}$ \quad
  Frank Hutter$^{1,2,5}$ \quad
  \\ \\
  $^1$University of Freiburg \quad
  $^2$ELLIS Institute Tübingen \quad \\
  $^3$Open-Sci Collective \quad
  $^4$LAION \quad
  $^5$Prior Labs \quad
}

\begin{document}

\maketitle

\begin{abstract}
    Symbolic regression (SR) aims to discover the underlying mathematical expressions that explain observed data. This holds promise for both gaining scientific insight and for producing inherently interpretable and generalizable models for tabular data. In this work we focus on the basics of SR. Deep learning-based SR has recently become competitive with genetic programming approaches, but the role of scale has remained largely unexplored. Inspired by scaling laws in language modeling, we present the first systematic investigation of scaling in SR, using a scalable end-to-end transformer pipeline and carefully generated training data. Across five different model sizes and spanning three orders of magnitude in compute, we find that both validation loss and solved rate follow clear power-law trends with compute. We further identify compute-optimal hyperparameter scaling: optimal batch size and learning rate grow with model size, and a token-to-parameter ratio of $\approx$15 is optimal in our regime, with a slight upward trend as compute increases. These results demonstrate that SR performance is largely predictable from compute and offer important insights for training the next generation of SR models.
\end{abstract}

\section{Introduction}

Symbolic regression seeks to uncover the underlying mathematical expressions that describe the relationship between a set of observed variables. In recent years, pre-trained transformer models for symbolic regression have become increasingly popular and have started to achieve performance comparable to classic genetic programming methods \citep{biggio2021neural, kamienny2022end, meidani2023snip, valipour2021symbolicgpt}. We observe that prior work has focused on tweaking the training process while holding the scale mostly constant. To date, we are not aware of any symbolic regression models trained with more than $\approx$100 million parameters.

Motivated by the impact of scaling laws in language modeling \citep{kaplan2020scaling, hoffmann2022training}, we ask: \emph{Do similar scaling laws exist for symbolic regression, and can they have a comparable impact on future model design?}
To answer this question, we see the need for a more controlled and scalable training setup. Specifically, we propose (1) a more systematic synthetic data generation approach that allows for stricter control over the expressions seen during training, and (2) targeted architectural improvements motivated by recent tabular foundation models. To keep the scope manageable yet insightful, our expressions contain only integer constants and at most two variables, while still covering a broad operator mix commonly seen in symbolic regression. Based on this setup, we conduct the first systematic scaling study in symbolic regression, spanning five model sizes and three orders of magnitude in compute. Our goal is not to beat existing baselines, but to establish that symbolic regression with transformers follows predictable scaling laws, which can serve as a design principle for future models.

Our key contributions are: (1) We demonstrate power-law scaling of solved rate and loss with compute over three orders of magnitude. (2) We identify systematic trends in optimal learning rate, batch size, and token-to-parameter ratio. (3) We introduce a scalable end-to-end pipeline that allows a clean and efficient scaling analysis.

\section{Related Work}

\citet{biggio2021neural} were the first to suggest pre-training transformers on millions of synthetic symbolic regression tasks. They use a set-encoder and a standard decoder to autoregressively predict function skeletons, whose constants are then refined via BFGS. Subsequent work has explored improvements on this theme, for example sharpening skeleton separation with a contrastive term \citep{li2022transformer} or framing the problem as multimodal translation \citep{meidani2023snip}. Other approaches, such as E2E \citep{kamienny2022end} and SymFormer \citep{vastl2024symformer} propose end-to-end architectures to directly output full expressions including constants, simplifying the pipeline and achieving comparable performance with genetic programming approaches.  
Our study builds on this end‑to‑end paradigm, coming closest to E2E, but shifts the focus from loss engineering to the role of scale.

\section{Methodology}

In this section, we describe our approach for data generation and model training. We adopt many ideas from \citet{kamienny2022end}, but apply targeted improvements that allow for more efficient training and a better scaling analysis.

\begin{figure}
  \centering
  \includegraphics[width=0.9\textwidth]{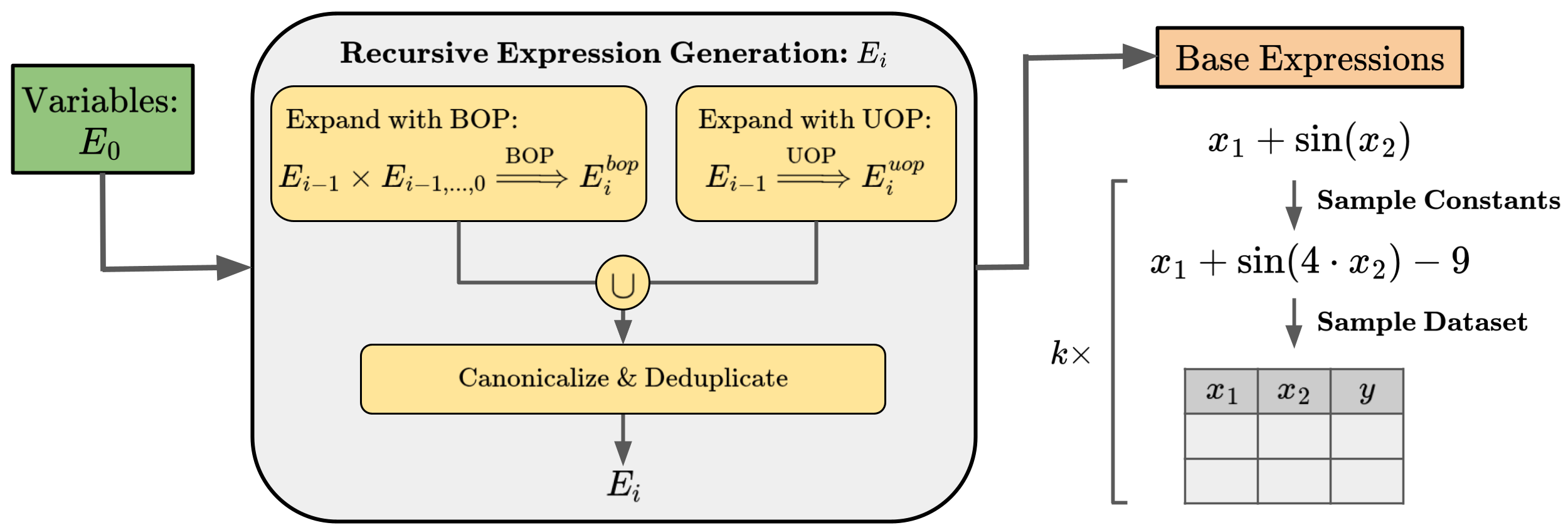}
  \caption{Overview of our two-step data generation. In the first step, we recursively generate a set of base expressions by applying a set of binary operators (BOP) and a set of unary operators (UOP). In the second step, we sample expression-dataset pairs from our base expressions.}
  \label{fig:data_generation}
\end{figure}

\textbf{Data generation}
Fig. \ref{fig:data_generation} shows our two-step approach for data generation. While previous work usually adopts the expression sampling mechanism proposed by \citet{lample2019deep}, we generate a complete set of expressions recursively, starting with just the variables and then iteratively applying unary and binary operators. This allows us to get more control over the expressions the model sees during training, especially in terms of the number of duplicates, and avoids biasing the model towards certain, more common expressions. Furthermore, we filter out equivalent expressions and unify their representation, allowing us to feed the model with cleaner training data. We obtain our set of base expressions $E$ by taking all expressions up to a fixed threshold. For each of these expressions, we try to sample $k$ expression-dataset pairs. Therefore, we first insert random constants and then sample a random dataset from a Gaussian mixture following \citet{kamienny2022end}. More details about our data generation can be found in App. \ref{app:data_details}.

\textbf{Tokenization}
Similar to existing approaches, we decide to use base 10 floating-point notation to encode the tabular dataset. Thus, we split each numeric value into a mantissa, including the sign, and an exponent. Under the assumption that different representations of expressions are equally difficult to learn for the model, we represent our target expressions as LaTeX strings for better readability. In the target expression, we tokenize all constants digit-wise.

\begin{figure}
  \centering
  \includegraphics[width=0.85\textwidth]{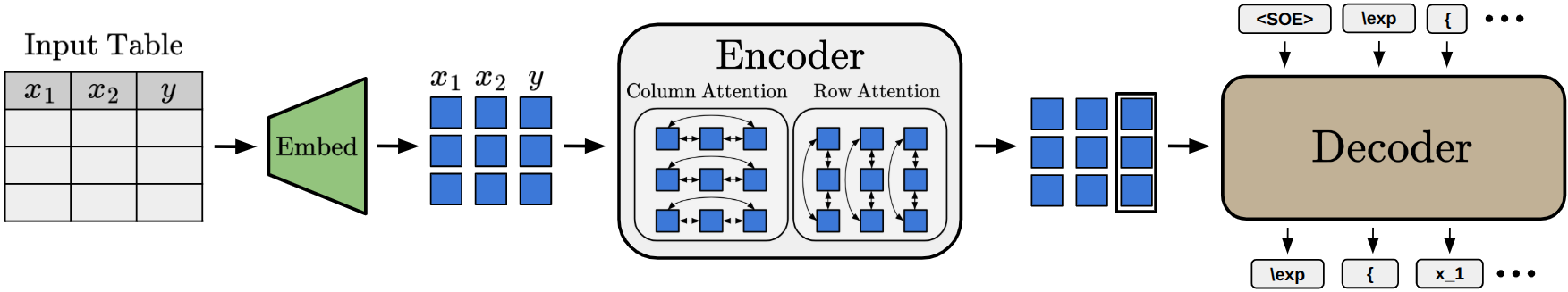}
  \caption{Overview of our model architecture.}
  \label{fig:model}
\end{figure}

\textbf{Architecture}
Given the great scaling properties of transformers \citep{raffel2020exploring, kaplan2020scaling}, we mainly adopt the standard encoder-decoder architecture used by previous work \citep{biggio2021neural, kamienny2022end} and only apply targeted improvements. Fig. \ref{fig:model} shows an overview of our architecture. While previous work decided to merge each input point into a single embedding, we choose a different embedding strategy and aim to get a single embedding for each cell in our dataset. Therefore, we up-project our mantissas and exponents to the embedding dimension and simply sum them together. This embedding strategy allows us to take inspiration from recent tabular foundation models and adopt their encoder architecture, which has proven to be very powerful for handling tabular data \citep{lorch2022amortized, hollmann2025accurate}. The main innovation here is that in each layer, we now apply both row- and column-wise attention across different variables and different data points. Our decoder follows the standard sequence-to-sequence transformer architecture \citep{vaswani2017attention} and only cross-attends to the updated embeddings of the target cells. We train with cross-entropy loss between the predicted tokens and the tokens of the true expression.

\textbf{Experiments}
To analyze scaling, we follow the strategies of scaling law papers in language modeling \citep{kaplan2020scaling, hoffmann2022training, porian2024resolving}. To approximate training compute, we adapt the FLOP estimate of \citet{kaplan2020scaling} to encoder-decoder models: FLOPs $\approx 6 \cdot (N_{enc} \cdot D_{in} + N_{dec} \cdot D_{out})$, with $N = N_{enc} + N_{dec}$ being the number of feed-forward parameters and $D_{in}$ and $D_{out}$ denoting the number of input and output tokens seen during training. For each of multiple model sizes, we conduct a sweep over different batch sizes and learning rates using a token-to-parameter ratio of $20$, which was found to be optimal for language model training \citep{hoffmann2022training}. We then retrain the models with the lowest validation loss per model size with varying token-to-parameter ratios.

\section{Experimental Results}

\subsection{Setup}

We use models of five different sizes (6.5M-93M parameters) and token-to-parameter ratios ranging from $5$ to $80$. For our training data, we generate $|E| = \num{100000}$ expressions with up to two variables. For each expression, we then sample up to $k = \num{3600}$ expression-dataset pairs by inserting random integer constants into the expression and sampling datasets of $64$ points. We also generate independent validation and test splits, each containing \num{1000} expressions sampled from $E$ but with freshly sampled constants and datasets. This ensures that no expression-dataset pair from training appears in evaluation.

We found that the AdamCPR optimizer \citep{franke2023constrained}, combined with a linear warm-up of the learning rate for the first 5\% of steps and cosine annealing thereafter, works best for our setting. All training hyperparameters are reported in App. \ref{app:training_details}. For evaluation, we first randomly sample $128$ expressions from the model and keep the one with the highest $R^2$ (coefficient of determination) score. Then, we look at both the perfect-solved ratio, denoted as $\text{Acc}_{\text{solved}}$, and the expression ratio with $R^2 > 0.99$, denoted as $\text{Acc}_{R^2>0.99}$, on the test split. We run each evaluation over three different seeds and report the mean $\text{Acc}_{\text{solved}}$ and $\text{Acc}_{R^2>0.99}$ across these seeds.

\subsection{Results}

We report the results of our hyperparameter sweep in App. \ref{app:additional_results} (Fig. \ref{fig:hparams_sweep}). Detailed evaluation results of the trainings with the best found hyperparameters can be found in App. \ref{app:additional_results} (Fig. \ref{tab:results_full}).
In the following, we highlight our three key findings.

\begin{figure}
  \centering
  \includegraphics[width=0.9\textwidth]{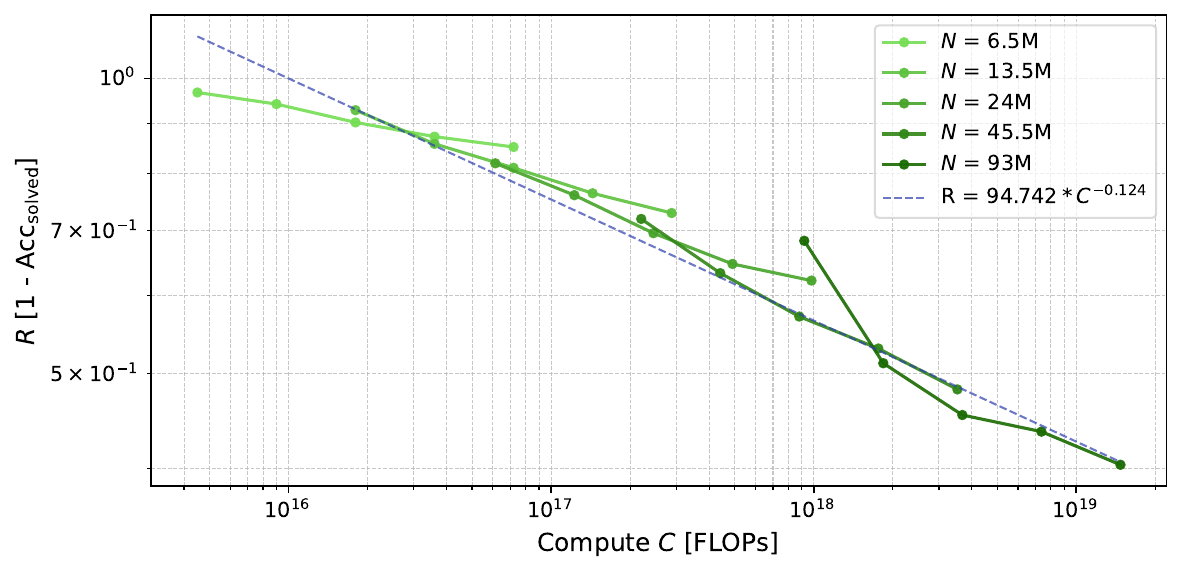}
  \vspace*{-0.2cm}
  \caption{$\text{Acc}_{\text{solved}}$ scales as a power law of training compute. Each marker corresponds to a trained model and depicts the mean perfect-solved ratio over three random seeds. Plot design inspired by \citet{franke2025learning}.}
  \label{fig:solved_rate_scaling}
  \vspace*{-0.2cm}
\end{figure}

\textbf{Solved rate and loss scale as power laws with compute}
To analyze the effect of scaling on performance, we look at training with the optimal configuration per model size. Following the approach of \citet{hoffmann2022training}, we get a Pareto front of our models by binning compute into \num{1500} intervals and keeping the models with the lowest validation loss up to their compute level. We adopt the assumption that scaling trends follow power-law trends and thus fit power laws on the remaining data. Fig. \ref{fig:solved_rate_scaling} shows the results for the $\text{Acc}_{\text{solved}}$ metric, which follows power-law scaling with compute across three orders of magnitude. In App. \ref{app:additional_results}, we report analogous plots for $\text{Acc}_{R^2>0.99}$ and test loss, yielding similar insights. Larger models with more data consistently improve the solved ratio and reduce loss, with no signs of saturation in the largest models tested. For instance, we can see that $\text{Acc}_{\text{solved}}$ increases from approximately $0.03$ at the lowest compute budget to $0.6$ at the highest compute budget, and that according to the scaling law, we could reach $0.8$ with a compute budget of $3.8 \times 10^{21}$ FLOPs. Also, by comparing the scaling laws, we can see that improvements in $\text{Acc}_{R^2>0.99}$ happen much faster than in $\text{Acc}_{\text{solved}}$, highlighting how hard it is to achieve exact expression matches.

\textbf{Optimal batch size and learning rate grow with compute}
Based on the results of our grid search, we analyze the scaling trends of the optimal learning rate and batch size. We follow the strategy of \citet{porian2024resolving}, which involves two-step interpolation by first interpolating the optimal learning rates and then the optimal batch sizes using Akima interpolation, followed by fitting scaling laws to both the optimal interpolated values. App. \ref{app:additional_results} (Fig. \ref{fig:hparams_scaling.pdf}) shows systematic trends for the compute-optimal choices of batch size and learning rate, indicating both should increase when scaling compute. The upward trend for the learning rate contrasts with findings for LLMs, showcasing how training characteristics can vary between different tasks. While we can see clear tendencies, we also note that more training runs over multiple seeds are needed to further reduce variance.

\textbf{Optimal trade-off between model size and data size} 
We fit scaling laws to the optimal number of parameters and expression tokens per compute budget. The results are shown in App. \ref{app:additional_results} (Fig. \ref{fig:optimal_trade_off.pdf}). We look at the ratio of the predicted scaling laws, noticing an increasing trend with growing compute. Although a token-to-parameter ratio of $\approx$15 seems to be optimal in our compute regime, the results indicate that the training dataset size should scale slightly faster than the model size.

\section{Limitations and Conclusion}

Our findings come with several limitations. We restrict the domain to expressions with at most two variables and small integer constants, whereas real-world symbolic regression tasks often involve more variables and floating-point constants. Thus, our predicted scaling laws might not directly translate to other symbolic regression setups. Due to computational constraints, we perform only single-seed training runs, which may introduce variance in our results. Moreover, our analysis spans a limited compute range, so extrapolations beyond this remain unverified. Finally, we do not compare against other symbolic regression methods, as we focus on general scaling insights, and comparison is difficult due to our customized setup.

Despite our limitations, our results show that symbolic regression with transformers follows predictable scaling laws. Performance improves with compute according to power laws, the optimal token-to-parameter ratio is approximately 15 for our compute budgets, and both batch size and learning rate should increase with model size. These insights provide practical heuristics for future work and suggest a new way of significantly improving model performance. Future work should extend our analysis to more complex expressions and verify that end-to-end symbolic regression with improved data generation, model architecture, and scaling can outperform all other approaches. We hope this work encourages a more systematic approach to scaling in symbolic regression.

\newpage

\section*{Acknowledgements}

The authors acknowledge support by the state of Baden-Württemberg through bwHPC and the German Research Foundation (DFG) through grant INST 35/1597-1 FUGG.
This research was funded by the Deutsche Forschungsgemeinschaft (DFG, German Research Foundation) under grant number 417962828.
We acknowledge funding by the European Union (via ERC Consolidator Grant DeepLearning 2.0, grant no.~101045765). Views and opinions expressed are, however, those of the author(s) only and do not necessarily reflect those of the European Union or the European Research Council. Neither the European Union nor the granting authority can be held responsible for them. 

\begin{center}\includegraphics[width=0.3\textwidth]{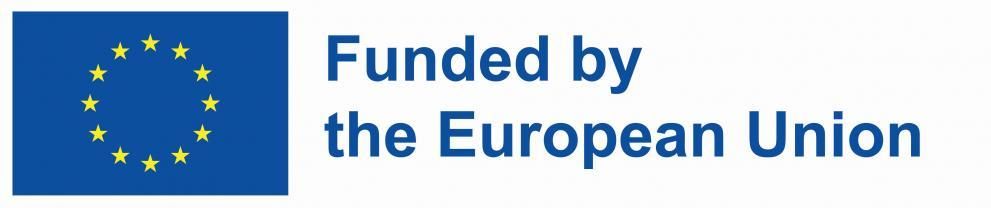}\end{center}

\bibliographystyle{unsrtnat}  
\bibliography{references}

@article{kamienny2022end,
  title={End-to-end symbolic regression with transformers},
  author={Kamienny, Pierre-Alexandre and d'Ascoli, St{\'e}phane and Lample, Guillaume and Charton, Fran{\c{c}}ois},
  journal={Advances in Neural Information Processing Systems},
  volume={35},
  pages={10269--10281},
  year={2022}
}

@inproceedings{biggio2021neural,
  title={Neural symbolic regression that scales},
  author={Biggio, Luca and Bendinelli, Tommaso and Neitz, Alexander and Lucchi, Aurelien and Parascandolo, Giambattista},
  booktitle={International Conference on Machine Learning},
  pages={936--945},
  year={2021},
  organization={Pmlr}
}

@article{kaplan2020scaling,
  title={Scaling laws for neural language models},
  author={Kaplan, Jared and McCandlish, Sam and Henighan, Tom and Brown, Tom B and Chess, Benjamin and Child, Rewon and Gray, Scott and Radford, Alec and Wu, Jeffrey and Amodei, Dario},
  journal={arXiv preprint arXiv:2001.08361},
  year={2020}
}

@article{hoffmann2022training,
  title={Training compute-optimal large language models},
  author={Hoffmann, Jordan and Borgeaud, Sebastian and Mensch, Arthur and Buchatskaya, Elena and Cai, Trevor and Rutherford, Eliza and Casas, Diego de Las and Hendricks, Lisa Anne and Welbl, Johannes and Clark, Aidan and others},
  journal={arXiv preprint arXiv:2203.15556},
  year={2022}
}

@article{porian2024resolving,
  title={Resolving discrepancies in compute-optimal scaling of language models},
  author={Porian, Tomer and Wortsman, Mitchell and Jitsev, Jenia and Schmidt, Ludwig and Carmon, Yair},
  journal={Advances in Neural Information Processing Systems},
  volume={37},
  pages={100535--100570},
  year={2024}
}

@article{lample2019deep,
  title={Deep learning for symbolic mathematics},
  author={Lample, Guillaume and Charton, Fran{\c{c}}ois},
  journal={arXiv preprint arXiv:1912.01412},
  year={2019}
}

@article{meidani2023snip,
  title={Snip: Bridging mathematical symbolic and numeric realms with unified pre-training},
  author={Meidani, Kazem and Shojaee, Parshin and Reddy, Chandan K and Farimani, Amir Barati},
  journal={arXiv preprint arXiv:2310.02227},
  year={2023}
}

@article{hollmann2025accurate,
  title={Accurate predictions on small data with a tabular foundation model},
  author={Hollmann, Noah and M{\"u}ller, Samuel and Purucker, Lennart and Krishnakumar, Arjun and K{\"o}rfer, Max and Hoo, Shi Bin and Schirrmeister, Robin Tibor and Hutter, Frank},
  journal={Nature},
  volume={637},
  number={8045},
  pages={319--326},
  year={2025},
  publisher={Nature Publishing Group UK London}
}

@article{lorch2022amortized,
  title={Amortized inference for causal structure learning},
  author={Lorch, Lars and Sussex, Scott and Rothfuss, Jonas and Krause, Andreas and Sch{\"o}lkopf, Bernhard},
  journal={Advances in Neural Information Processing Systems},
  volume={35},
  pages={13104--13118},
  year={2022}
}

@article{franke2023constrained,
  title={Constrained parameter regularization},
  author={Franke, J{\"o}rg KH and Hefenbrock, Michael and Koehler, Gregor and Hutter, Frank},
  year={2023}
}

@inproceedings{li2022transformer,
  title={Transformer-based model for symbolic regression via joint supervised learning},
  author={Li, Wenqiang and Li, Weijun and Sun, Linjun and Wu, Min and Yu, Lina and Liu, Jingyi and Li, Yanjie and Tian, Songsong},
  booktitle={The Eleventh International Conference on Learning Representations},
  year={2022}
}

@article{vastl2024symformer,
  title={Symformer: End-to-end symbolic regression using transformer-based architecture},
  author={Vastl, Martin and Kulh{\'a}nek, Jon{\'a}{\v{s}} and Kubal{\'\i}k, Ji{\v{r}}{\'\i} and Derner, Erik and Babu{\v{s}}ka, Robert},
  journal={IEEE Access},
  volume={12},
  pages={37840--37849},
  year={2024},
  publisher={IEEE}
}

@article{vaswani2017attention,
  title={Attention is all you need},
  author={Vaswani, Ashish and Shazeer, Noam and Parmar, Niki and Uszkoreit, Jakob and Jones, Llion and Gomez, Aidan N and Kaiser, {\L}ukasz and Polosukhin, Illia},
  journal={Advances in neural information processing systems},
  volume={30},
  year={2017}
}

@article{franke2025learning,
  title={Learning in Compact Spaces with Approximately Normalized Transformers},
  author={Franke, J{\"o}rg KH and Spiegelhalter, Urs and Nezhurina, Marianna and Jitsev, Jenia and Hutter, Frank and Hefenbrock, Michael},
  journal={arXiv preprint arXiv:2505.22014},
  year={2025}
}

@article{wortsman2023small,
  title={Small-scale proxies for large-scale transformer training instabilities},
  author={Wortsman, Mitchell and Liu, Peter J and Xiao, Lechao and Everett, Katie and Alemi, Alex and Adlam, Ben and Co-Reyes, John D and Gur, Izzeddin and Kumar, Abhishek and Novak, Roman and others},
  journal={arXiv preprint arXiv:2309.14322},
  year={2023}
}

@article{meurer2017sympy,
  title={SymPy: symbolic computing in Python},
  author={Meurer, Aaron and Smith, Christopher P and Paprocki, Mateusz and {\v{C}}ert{\'\i}k, Ond{\v{r}}ej and Kirpichev, Sergey B and Rocklin, Matthew and Kumar, AMiT and Ivanov, Sergiu and Moore, Jason K and Singh, Sartaj and others},
  journal={PeerJ Computer Science},
  volume={3},
  pages={e103},
  year={2017},
  publisher={PeerJ Inc.}
}

@article{valipour2021symbolicgpt,
  title={Symbolicgpt: A generative transformer model for symbolic regression},
  author={Valipour, Mojtaba and You, Bowen and Panju, Maysum and Ghodsi, Ali},
  journal={arXiv preprint arXiv:2106.14131},
  year={2021}
}

@article{raffel2020exploring,
  title={Exploring the limits of transfer learning with a unified text-to-text transformer},
  author={Raffel, Colin and Shazeer, Noam and Roberts, Adam and Lee, Katherine and Narang, Sharan and Matena, Michael and Zhou, Yanqi and Li, Wei and Liu, Peter J},
  journal={Journal of machine learning research},
  volume={21},
  number={140},
  pages={1--67},
  year={2020}
}


\newpage
\appendix

\section{Statement of contribution}

DO developed the data generation methodology, implemented all software components, conducted the formal analysis and all experiments, curated the datasets, and created the visualizations. 
JF conceptualised the main research idea and designed the model architecture and scaling analysis methodology. 
DO wrote the original draft and JF and DO performed review and editing. 
JF and FH supervised the project and coordinated the research activities. 
FF and AZ participated on preliminary discussions.
FH acquired funding and provided computing resources.

\section{Data generation details}
\label{app:data_details}

\subsection{Expression generation}

\paragraph{Recursive generation} Our general goal is to get a complete set of unique symbolic expressions up to a certain size. We try to approach this by looking at the depth of the corresponding expression trees. We start with all expressions of depth 0, which are just our variables: $E_0 = \{x_1, ..., x_n\}$. Given all expressions up to a certain depth, we can get all possible expressions of the subsequent depth by: (1) Applying all unary operators to all expressions in $E_{i-1}$ and (2) applying all binary operators to every pair of expressions in $E_{i-1}$ and $E_{i-1,...,0} = E_{i-1} \cup \dots \cup E_0$.

\paragraph{Canonicalization and deduplication} For having consistent training data and avoiding duplicates, we try to bring every generated expression into a canonical form using Sympy \citep{meurer2017sympy}. We avoid using their simplify function, as it is based on heuristics and gives inconsistent outputs. Instead, we expand all expressions and iteratively check for specific simplifications we can do. Finally, we apply a custom ordering function. For deduplication, we can now simply check if a new expression was already found before and add it to our set of new expressions.

\subsection{Data sampling}

We continue with the first base expressions of $E_0, ..., E_d$ up to a certain threshold. This means taking the full generated sets of lower depths and sampling expressions of the final depth we want to consider until we reach the threshold. Now, for each of the base expressions we try to sample $k$ expression-dataset pairs. For each sampled pair, we first insert random constants into the base expression, by sampling a multiplicative and an additive constant for each variable and each unary operator by a probability $p$. Then, we sample a dataset following \citet{kamienny2022end}, with retrying up to five times in case of failure. Specifically, this approach involves: (1) Sampling the number of clusters and weights for each cluster. (2) Sampling centroids, variances and a distribution shape (Gaussian or uniform) for each cluster. (3) Sampling a dataset and applying a randomly sampled rotation from the Haar distribution.

\begin{table}[h!]
  \centering
  \caption{Used hyperparameters for data generation.}
  \begin{tabular}{|c|c|}
    \hline
    \textbf{Parameter} & \textbf{Value} \\
    \hline
    Number of variables & 2 \\
    Max tree depth & 3 \\
    BOP & \{+, -, $\cdot$, $\div$\} \\
    UOP & \{exp, sin, neg, sqrt\} \\
    $|E|$ & \num{100000} \\
    $k$ & \num{3600} \\
    Data points per dataset & 64 \\
    Constant type & integer \\
    Constant range & -9 to 9 \\
    Constant probability $p$ & 0.2 \\
    Max number of clusters & 5 \\
    Dataset sampling retries & 5 \\
    \hline
  \end{tabular}
  \label{tab:data_hparams}
\end{table}

\newpage

\section{Additional Results}
\label{app:additional_results}

\begin{figure}[h!]
  \centering
  \includegraphics[width=0.9\textwidth]{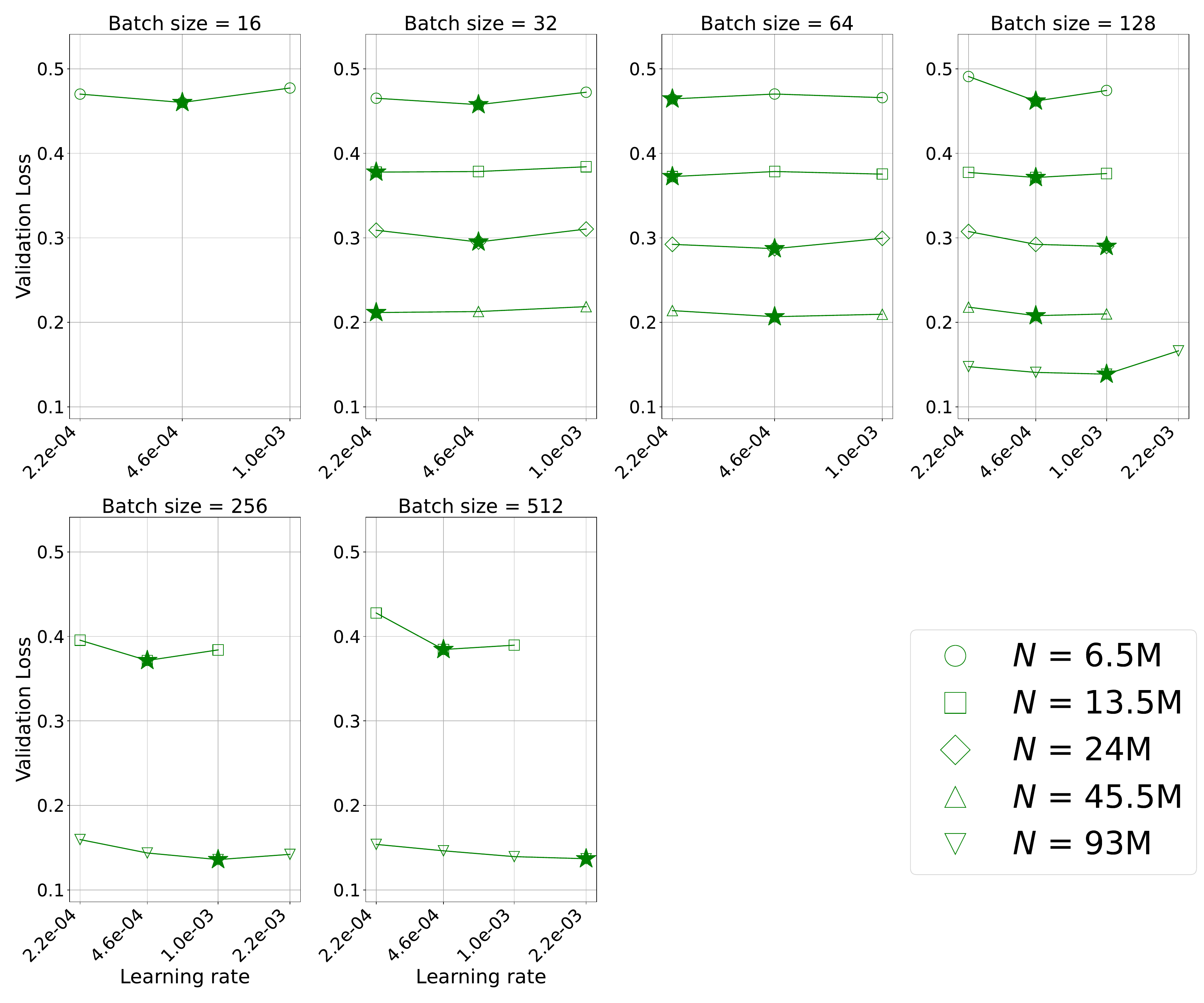}
  \caption{Results of our hyperparameter grid search. For each model size we trained different combinations of batch size and learning rate using a token-to-parameter ratio of 20, until we found an optimum. The stars indicate the runs with the lowest loss for each model size and batch size. Plot design inspired by \citet{wortsman2023small}.}
  \label{fig:hparams_sweep}
\end{figure}

\begin{figure}[h!]
  \centering
  \includegraphics[width=1.0\textwidth]{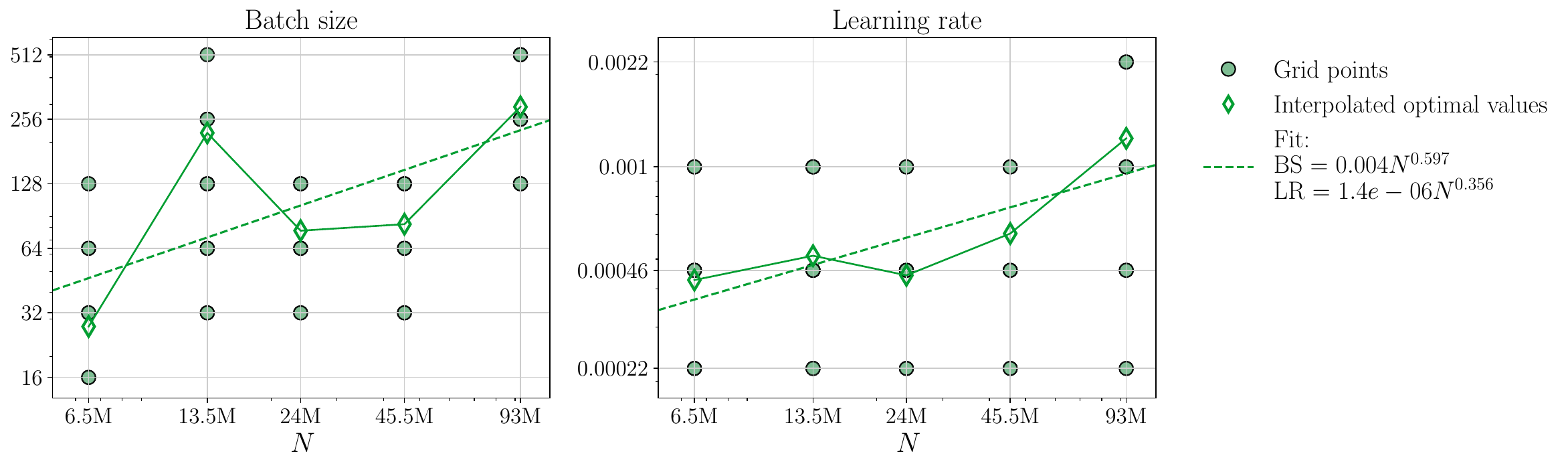}
  \caption{Compute-optimal batch size and learning rate as functions of model size $N$.}
  \label{fig:hparams_scaling.pdf}
\end{figure}

\begin{figure}[h!]
  \centering
  \begin{subfigure}{0.48\textwidth}
    \centering
    \includegraphics[width=\linewidth]{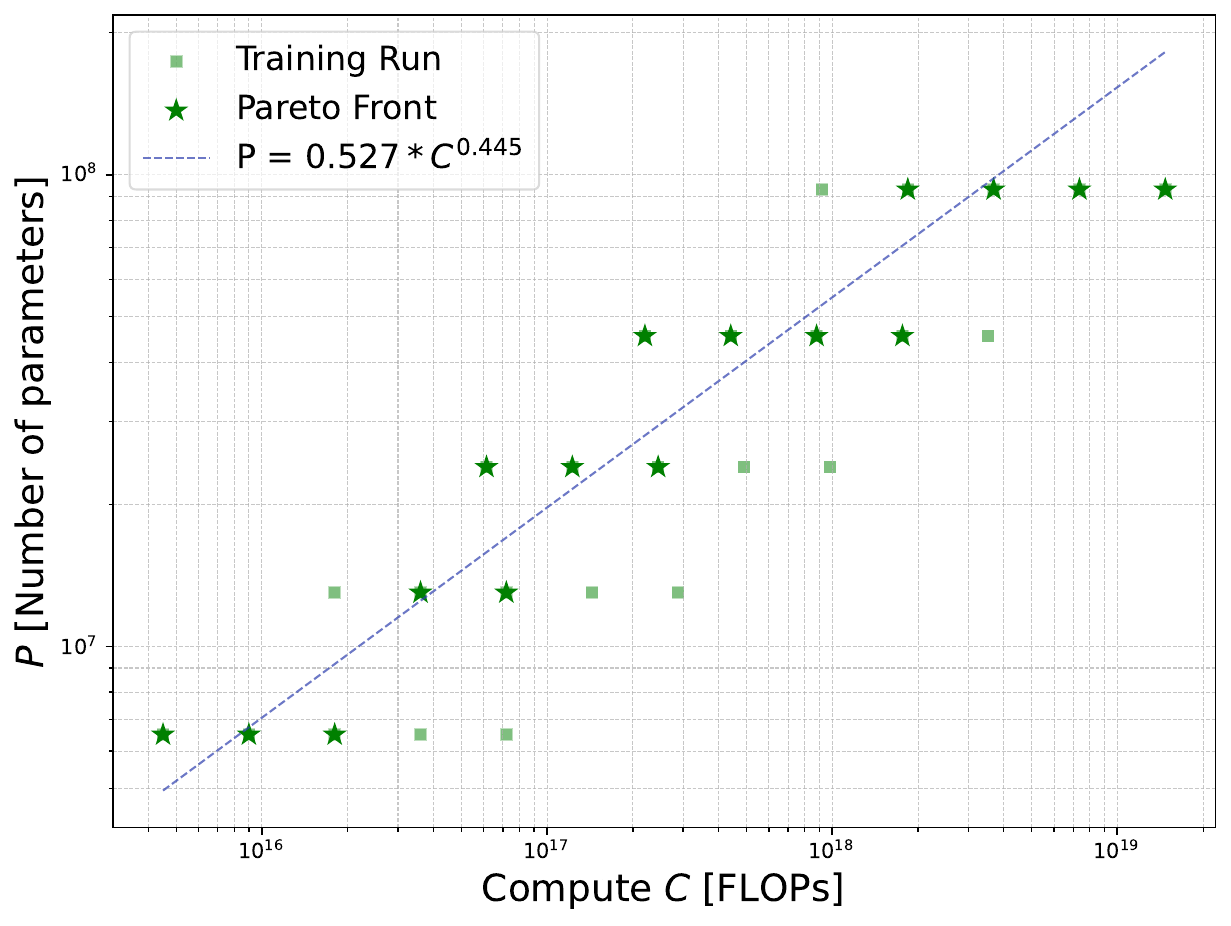}
    \caption{Compute-optimal number of parameters}
    \label{fig:img1}
  \end{subfigure}
  \hfill
  \begin{subfigure}{0.48\textwidth}
    \centering
    \includegraphics[width=\linewidth]{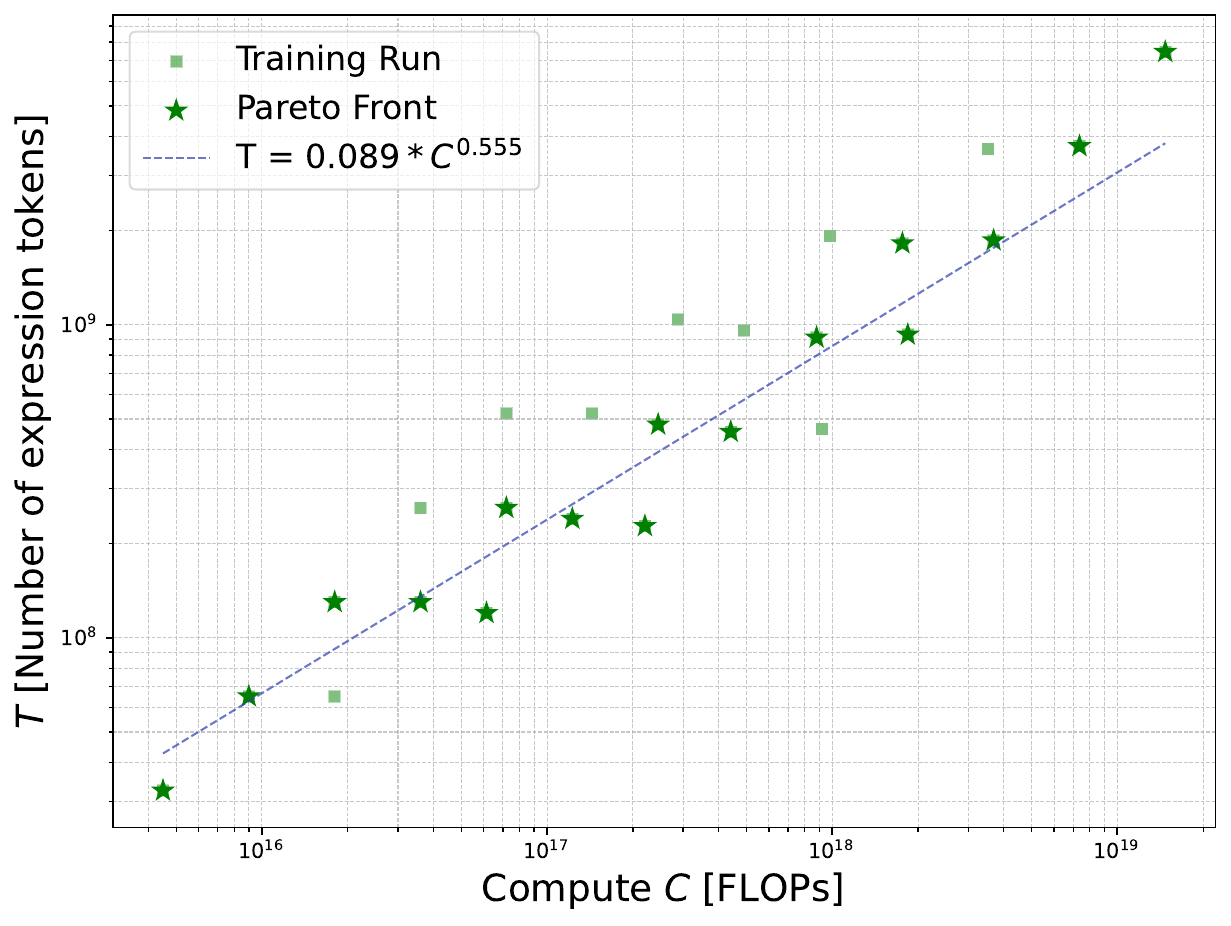}
    \caption{Compute-optimal dataset size}
    \label{fig:img2}
  \end{subfigure}
  \caption{Both optimal number of parameters and training tokens scale as power laws with compute.}
  \label{fig:optimal_trade_off.pdf}
\end{figure}

\begin{figure}[h!]
  \centering
  \includegraphics[width=0.9\textwidth]{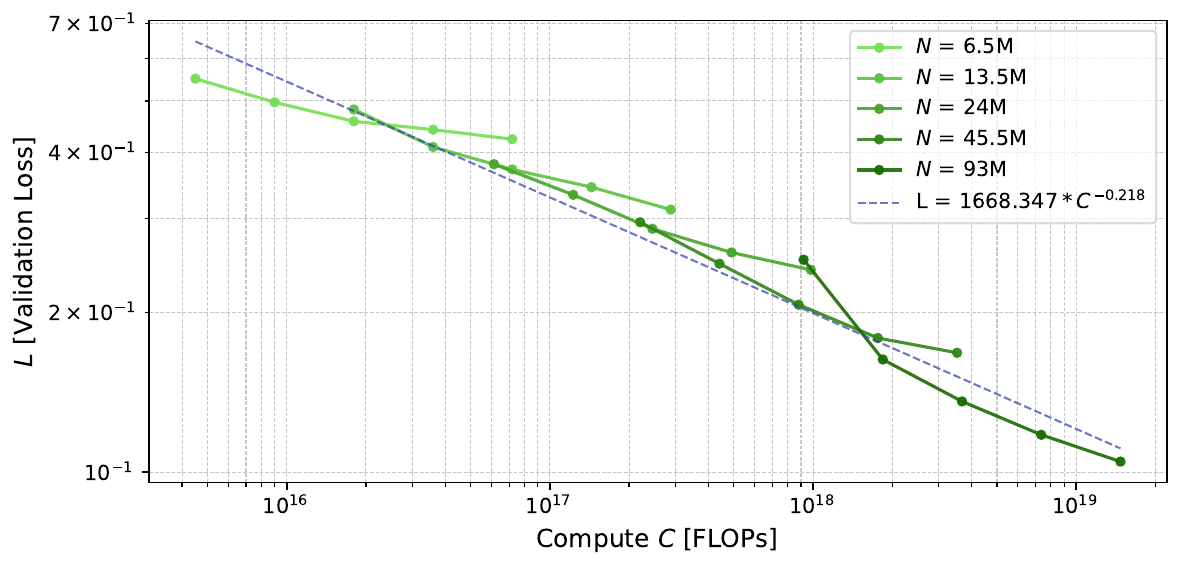}
  \caption{Validation loss scales as a power law of training compute. Each marker corresponds to a trained model and depicts the final validation loss after training.}
  \label{fig:loss_scaling}
\end{figure}

\begin{figure}[h!]
  \centering
  \includegraphics[width=0.9\textwidth]{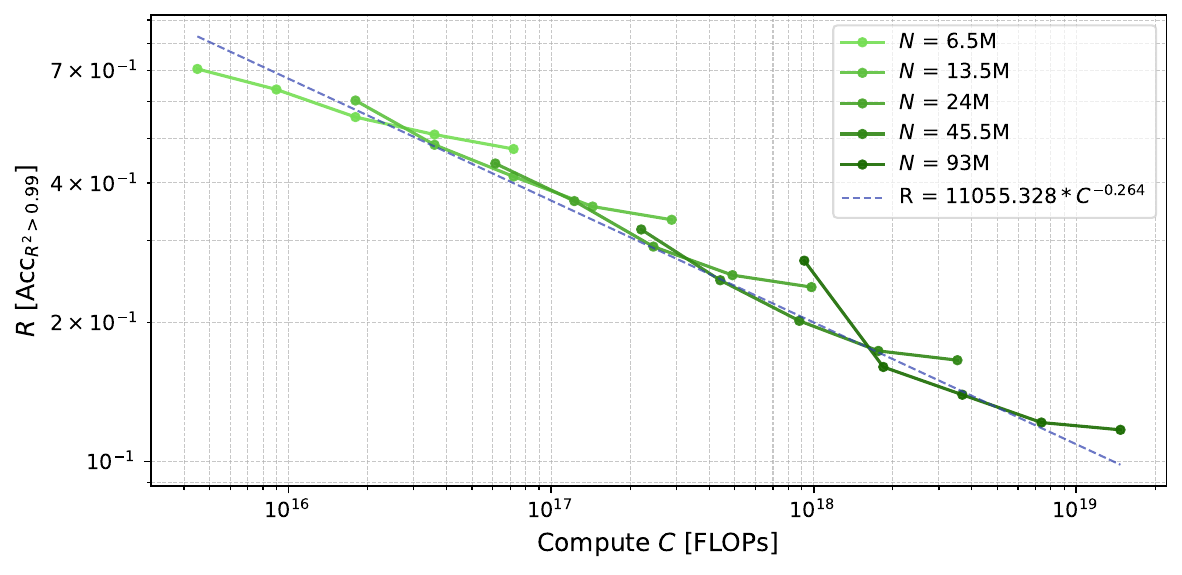}
  \caption{$\text{Acc}_{R^2>0.99}$ scales as a power law of training compute. Each marker corresponds to a trained model and depicts the mean $\text{Acc}_{R^2>0.99}$ over three random seeds.}
  \label{fig:accuracy_scaling}
\end{figure}

\begin{table}[htbp]
  \centering
  \caption{Full result table of runs with best found hyperparameters.}
  \resizebox{\textwidth}{!}{
  \begin{tabular}{|c|ccc|ccc|}
    \hline
    \textbf{Model Size} & \textbf{Batch Size} & \textbf{Learning Rate} & \textbf{Training FLOPs} & \textbf{$\text{Acc}_{\text{solved}}$} & \textbf{$\text{Acc}_{R^2>0.99}$} & \textbf{Final Validation Loss} \\
    \hline
    6.5M & 32 & 4.6e-4 & 4.50e+15 & 0.0327 & 0.2943 & 0.5506 \\
    6.5M & 32 & 4.6e-4 & 8.99e+15 & 0.0587 & 0.3630 & 0.4969 \\
    6.5M & 32 & 4.6e-4 & 1.80e+16 & 0.0983 & 0.4447 & 0.4576 \\
    6.5M & 32 & 4.6e-4 & 3.60e+16 & 0.1280 & 0.4910 & 0.4414 \\
    6.5M & 32 & 4.6e-4 & 7.20e+16 & 0.1490 & 0.5263 & 0.4235 \\
    13.5M & 128 & 4.6e-4 & 1.80e+16 & 0.0720 & 0.3970 & 0.4818 \\
    13.5M & 128 & 4.6e-4 & 3.60e+16 & 0.1427 & 0.5163 & 0.4095 \\
    13.5M & 128 & 4.6e-4 & 7.20e+16 & 0.1897 & 0.5870 & 0.3714 \\
    13.5M & 128 & 4.6e-4 & 1.44e+17 & 0.2367 & 0.6443 & 0.3440 \\
    13.5M & 128 & 4.6e-4 & 2.88e+17 & 0.2713 & 0.6670 & 0.3121 \\
    24M & 64 & 4.6e-4 & 6.13e+16 & 0.1810 & 0.5593 & 0.3800 \\
    24M & 64 & 4.6e-4 & 1.23e+17 & 0.2401 & 0.6345 & 0.3328 \\
    24M & 64 & 4.6e-4 & 2.45e+17 & 0.3050 & 0.7087 & 0.2872 \\
    24M & 64 & 4.6e-4 & 4.91e+17 & 0.3535 & 0.7473 & 0.2591 \\
    24M & 64 & 4.6e-4 & 9.81e+17 & 0.3783 & 0.7619 & 0.2404 \\
    45.5M & 64 & 4.6e-4 & 2.20e+17 & 0.2813 & 0.6827 & 0.2955 \\
    45.5M & 64 & 4.6e-4 & 4.41e+17 & 0.3673 & 0.7536 & 0.2469 \\
    45.5M & 64 & 4.6e-4 & 8.81e+17 & 0.4287 & 0.7987 & 0.2067 \\
    45.5M & 64 & 4.6e-4 & 1.76e+18 & 0.4700 & 0.8267 & 0.1789 \\
    45.5M & 64 & 4.6e-4 & 3.53e+18 & 0.5185 & 0.8345 & 0.1678 \\
    93M & 256 & 1.0e-3 & 9.21e+17 & 0.3173 & 0.7283 & 0.2512 \\
    93M & 256 & 1.0e-3 & 1.84e+18 & 0.4880 & 0.8400 & 0.1630 \\
    93M & 256 & 1.0e-3 & 3.68e+18 & 0.5467 & 0.8607 & 0.1359 \\
    93M & 256 & 1.0e-3 & 7.37e+18 & 0.5640 & 0.8787 & 0.1176 \\
    93M & 256 & 1.0e-3 & 1.47e+19 & 0.5967 & 0.8830 & 0.1047 \\
    \hline
  \end{tabular}
  }
  \label{tab:results_full}
\end{table}

\FloatBarrier

\section{Training Details}
\label{app:training_details}

\begin{table}[h!]
  \centering
  \caption{Architectural details of our five different model sizes.}
  \begin{tabular}{|c|ccccc|}
    \hline
    \textbf{Model} & \textbf{6.5M} & \textbf{13.5M} & \textbf{24M} & \textbf{45.5M} & \textbf{93M}\\
    \hline
    Model dimension & 256 & 320 & 384 & 448 & 512 \\
    Number of encoder layers & 3 & 4 & 5 & 7 & 11 \\
    Number of decoder layers & 3 & 4 & 5 & 7 & 11 \\
    Number of attention heads & 4 & 5 & 6 & 7 & 8 \\
    Head dimension & 64 & 64 & 64 & 64 & 64 \\
    MLP dimension & \num{1024} & \num{1280} & \num{1536} & \num{1792} & \num{2048} \\
    \hline
    Parameters & 6.48M & 13.40M & 24.01M & 45.53M & 93.08M \\
    \hline
  \end{tabular}
  \label{tab:model_details}
\end{table}

\begin{table}[h!]
  \centering
  \caption{List of training hyperparameters.}
  \begin{tabular}{|c|c|}
    \hline
    \textbf{Parameter} & \textbf{Value}\\
    \hline
    Clip value & 1.0 \\
    Weight decay & 0.1 \\
    Optimizer & AdamCPR \\
    Beta1 & 0.9 \\
    Beta2 & 0.98 \\
    Eps & 1.0e-09 \\
    Kappa init param & \num{1000} \\
    Kappa init method & inflection\_point \\
    Reg function & l2 \\
    Kappa update & 1.0 \\
    LR warmup steps & 5\% \\
    LR decay factor & 0.01 \\
    LR schedule & cosine \\
    Residual dropout & 0.1 \\
    Attention dropout & 0.1 \\
    LN eps & 1e-5 \\
    Param init scale & 0.02 \\
    Precision & fp32 \\
    Max Output Length & 256 \\
    \hline
  \end{tabular}
  \label{tab:train_hparams}
\end{table}


\end{document}